\documentclass[letterpaper, 10 pt, conference]{ieeeconf}  

\IEEEoverridecommandlockouts                              

\usepackage{graphicx}
\usepackage{amsmath}
\usepackage{amssymb}
\usepackage{url}
\usepackage{hyperref}

\usepackage{color, colortbl}
\usepackage{array,multirow,graphicx}
\usepackage{siunitx}

\graphicspath{ {imgs/} }
\usepackage{tikz}
\usepackage{epstopdf}
\usepackage{caption}
\usepackage{subcaption}

\definecolor{Gray}{gray}{0.9}
\definecolor{Gray2}{rgb}{0.1,0.1,0.1}

\newif\ifshowComments
\showCommentstrue   

\ifshowComments
\usepackage[draft,markup=bfit, deletedmarkup=sout, authormarkup=brackets]{changes}
\else
\usepackage[final,markup=bfit, deletedmarkup=sout, authormarkup=brackets]{changes}
\fi

\usepackage{draftwatermark}

\colorlet{Changes@Color}{blue}
\definechangesauthor[name={Honza}, color=red]{JD}
\definechangesauthor[name={Matej}, color=orange]{MH}
\definechangesauthor[name={Kuba}, color=green]{JR}
\definechangesauthor[name={Jiri}, color=magenta]{JM}

\ifshowComments

\else

\fi

\newcommand{\figref}[1]{Fig.\@~\ref{#1}}
\newcommand{\tabref}[1]{Tab.\@~\ref{#1}}

\title{Human keypoint detection for close proximity human-robot interaction}

\author{Jan Docekal, Jakub Rozlivek, Jiri Matas, and Matej Hoffmann
\thanks{This work was supported by the OP VVV MEYS funded project CZ.02.1.01/0.0/0.0/16\_019/0000765 ``Research Center for Informatics''. J.R. was additionally supported by the Czech Technical University in Prague, grant no. SGS22/111/OHK3/2T/13.}
\thanks{Jan Docekal, Jakub Rozlivek, Jiri Matas, and Matej Hoffmann are with Department of Cybernetics, Faculty of Electrical Engineering, Czech Technical University in Prague, Czech Republic. {\tt\footnotesize matej.hoffmann@fel.cvut.cz}.}%
}

\begin{document}

\SetWatermarkAngle{0}
\SetWatermarkColor{black}
\SetWatermarkLightness{0.5}
\SetWatermarkFontSize{10pt}
\SetWatermarkVerCenter{25pt}
\SetWatermarkText{\parbox{30cm}{%
\centering This is the authors' final version of the manuscript published as\\
\centering Docekal, J.; Rozlivek, J.; Matas, J. \& Hoffmann, M. (2022), Human keypoint detection for close proximity human-robot interaction \\
\centering  in 'IEEE-RAS International Conference on Humanoid Robots (Humanoids 2022)', pp. 450-457. (C) IEEE \\
\centering \url{https://doi.org/10.1109/Humanoids53995.2022.10000133}
}}

\maketitle

\begin{abstract}
We study the performance of state-of-the-art human keypoint detectors in the context of close proximity human-robot interaction.
The detection in this scenario is specific in that  only a subset of body parts such as hands and torso are in the field of view. 
In particular, 
(i) we survey existing datasets with human pose annotation from the perspective of close proximity images and prepare and make publicly available a new Human in Close Proximity (HiCP) dataset;
(ii) we quantitatively and qualitatively compare state-of-the-art human whole-body 2D keypoint detection methods (OpenPose, MMPose, AlphaPose, Detectron2) on this dataset; (iii) since accurate detection of hands and fingers is critical in applications with handovers, we evaluate the performance of the MediaPipe hand detector; (iv) we deploy the algorithms on a humanoid robot with an RGB-D camera on its head and evaluate the performance in 3D human keypoint detection. A motion capture system is used as reference.

The best performing whole-body keypoint detectors in close proximity were  MMPose and AlphaPose, but both had difficulty with finger detection. Thus, we propose a combination of MMPose or AlphaPose for the body and MediaPipe for the hands in a single framework providing the most accurate and robust detection. We also analyse the failure modes of individual detectors---for example, to what extent the absence of the head of the person in the image degrades performance. Finally, we demonstrate the framework in a scenario where a humanoid robot interacting with a person uses the detected 3D keypoints for whole-body avoidance maneuvers.

\end{abstract}

\section{Introduction}
\begin{figure}[tb]
    \centering
    \includegraphics[width=0.275\textwidth]{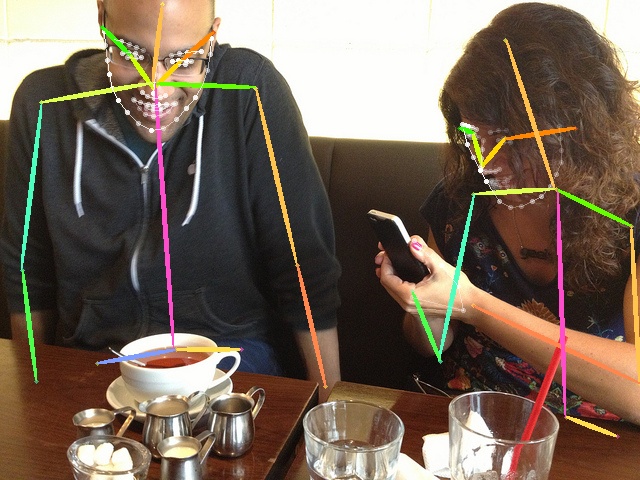}
    \includegraphics[width=0.2\textwidth]{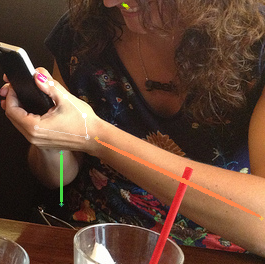}
    \caption{Human keypoints detected by the AlphaPose detector.  An image from the Halpe dataset (left),  a close proximity simulated image by cropping, from the HiCP dataset (right). Note the missing keypoints on the neck and shoulder in the proximity image (yellow and green segments on the left).}
    \label{fig:motivation}
    \vspace*{-6mm}
\end{figure}

Detection of humans and their pose in images or videos is useful in various applications. Recently, a number of solutions based on deep convolutional neural networks have appeared that recognize human poses in images. In human-robot interaction (HRI), human pose and keypoint detection is important both for social interaction (to recognize gestures or to keep socially acceptable interpersonal distance) and physical interaction. For the latter, accurate and reliable detection of human body parts is key for safety. Relevant safety standards~\cite{ISO/TS15066} prescribe specific regimes for warranting safe interaction, such as speed and separation monitoring, whereby the distance between the robot and the human is gauged and it needs to be guaranteed that the machine can stop before colliding with the operator.  Solutions in industry relying on 2D or 3D scanners are inflexible and result in large protective zones and corresponding footprint of the robot on the factory floor. Detecting individual parts of the human body and their current position with respect to the moving robot brings higher resolution, flexibility, and potential to boost productivity of a collaborative application (e.g., \cite{Svarny_SSR_2018,svarny2019}). However, this scenario differs from the standard case, whereby complete human bodies from considerable distances are detected in images by pose detectors. For physical HRI---in particular for humanoid robots with sensors typically on the body---the human is seen from close proximity and hence only some body parts are visible (see \figref{fig:motivation}). Second, pixel coordinates of the human body in the image are not sufficient. This is where this work ties in: we assess the performance of several state-of-the-art human pose detectors in the close proximity scenario, including their conversion from 2D to 3D.

The main contributions of this paper are the following: (i) survey of existing datasets with human pose annotation and creating a new Human in Close Proximity (HiCP) dataset; (ii) quantitative and qualitative comparison of state-of-the-art human whole-body keypoint detection methods (OpenPose, MMPose, AlphaPose, Detectron2) and special hand-only detector (MediaPipe) on this dataset; (iii) experimental evaluation using an RGB-D camera on the moving head of a humanoid robot and assessment of keypoint detection in 3D; (iv) deployment in a closed-loop scenario with a humanoid robot monitoring a human and performing real-time whole-body avoidance.


This article is structured as follows. After we review related work, in Section~\ref{sec:met}, we present our new close proximity keypoint dataset, methods relevant for the evaluation of experiments, and robot experiments. Then we present experimental results and wrap up with Conclusion, Discussion, and Future work. 

\section{Related work}
\label{sec:related}

\subsection{Human keypoints datasets}
\begin{figure}[tb]
    \centering
    \begin{subfigure}[t]{.235\textwidth}
    \centering
    \includegraphics[height=0.1\textheight]{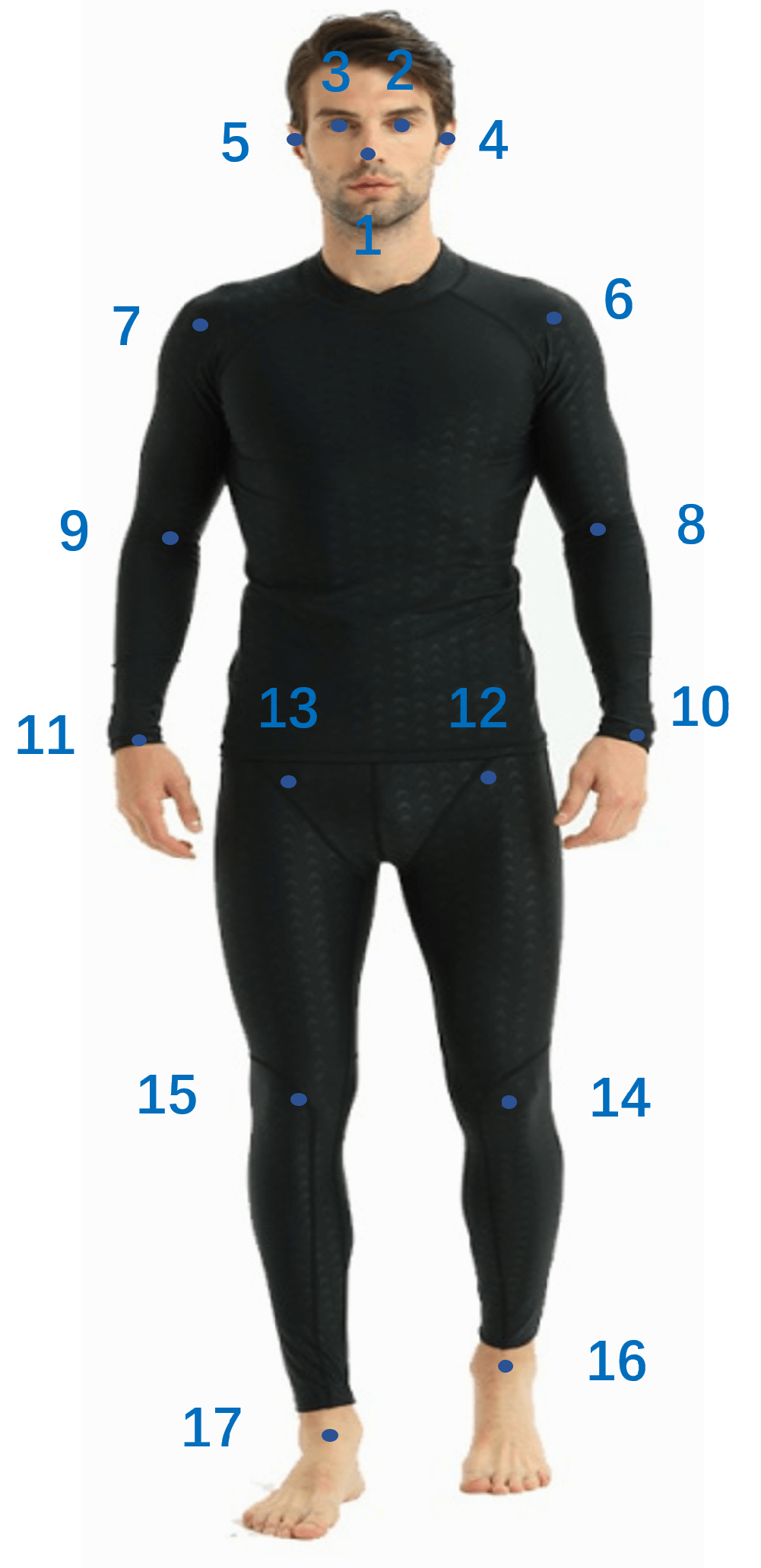}
    \caption{\small 17 (COCO) \texttt{body} keypoints.}
    \label{fig:body17}
    \end{subfigure}
    \begin{subfigure}[t]{.24\textwidth}
    \centering
    \includegraphics[height=0.1\textheight]{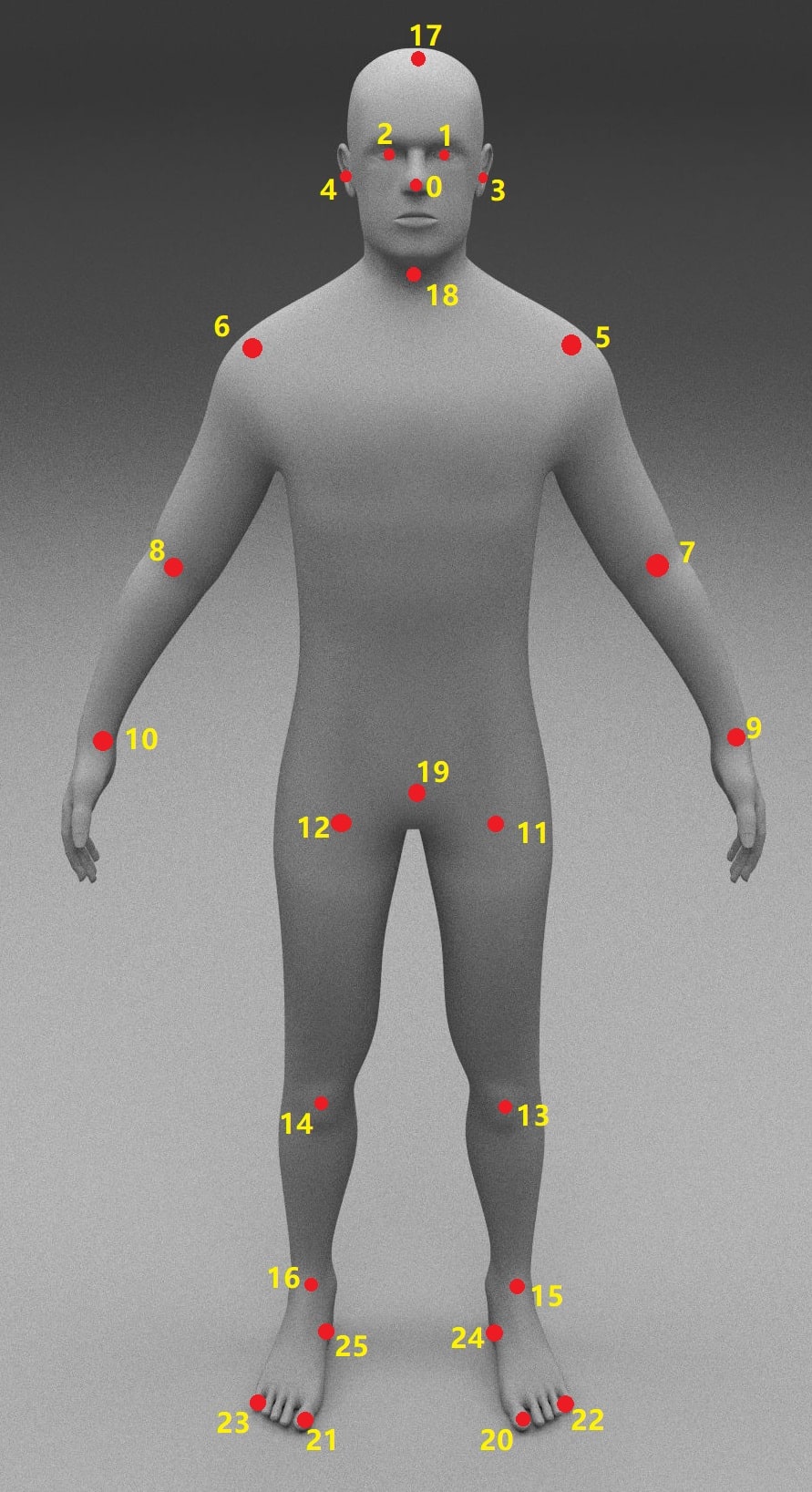}
    \caption{\small 26 (Halpe) \texttt{body} keypoints.}
    \label{fig:body26}
    \end{subfigure}
    \begin{subfigure}[t]{0.235\textwidth}
    \centering
    \includegraphics[height=0.1\textheight]{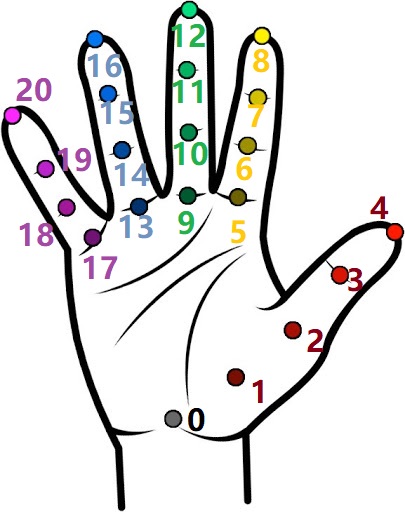}
    \caption{\small 21 \texttt{hand} keypoints.}
    \label{fig:hand_kpts}
    \end{subfigure}%
    \begin{subfigure}[t]{.24\textwidth}
    \centering
    \includegraphics[height=0.1\textheight]{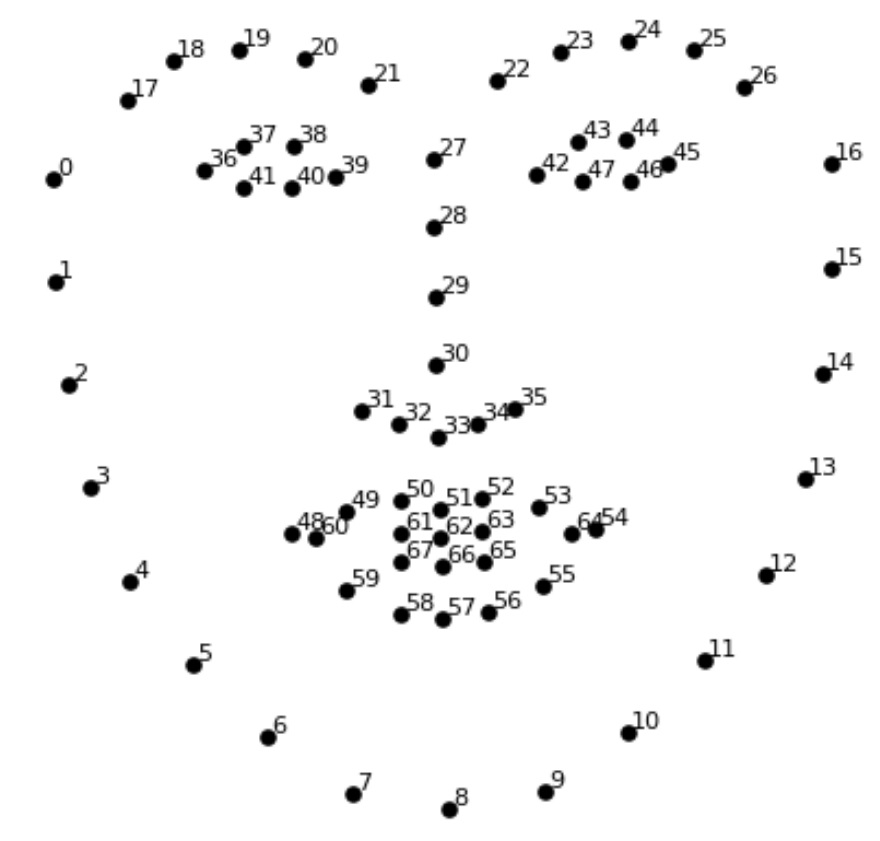}
    \caption{\small 68 \texttt{face} keypoints.}
    \label{fig:face_kpts}
    \end{subfigure}
    \caption{\small Selected human body keypoint standards.\protect\footnotemark}
    \label{fig:keypoints_standard}
    \vspace*{-5mm}
\end{figure}

Human keypoints can be divided into three main groups. 1) \texttt{body} keypoints containing basic keypoints of arms, legs, torso, and head; two examples with 17 and 26 keypoints are shown in Figs.\@~\ref{fig:body17} and \ref{fig:body26}, respectively; 2)  \texttt{hand} keypoints containing 21 keypoints in a palm (see \figref{fig:hand_kpts}); 3) \texttt{face} keypoints covering a head with 68 keypoints (see \figref{fig:face_kpts}). 

\footnotetext{Image \ref{fig:body17} taken from \cite{jin2020whole}. Images \ref{fig:body26}, \ref{fig:hand_kpts}, and \ref{fig:face_kpts} taken from \url{https://github.com/Fang-Haoshu/Halpe-FullBody}.}

There are several well-known datasets with annotated human keypoints. The first of them is the COCO dataset \cite{coco_dataset}. It consists of more than 200,000 images with several features, such as object segmentation, object detection, or people with annotated keypoints. 
The original human keypoint annotations provided with the COCO dataset have 17 keypoints per person. Subsequently, these were extended to COCO-WholeBody annotations \cite{jin2020whole} with finger, foot, and face keypoints, resulting in 133 keypoints together. The resulting dataset contains about 64,000 training images and 2,700 validation images with people in the scene.

The Halpe dataset \cite{fang2017rmpe, li2020pastanet} contains human keypoints similar to the COCO-WholeBody annotations, with slight differences in the number of keypoints (136) and annotation format. 
It has 41,000 training images with people and shares the same validation images with the COCO dataset. 

The DensePose-Posetrack dataset \cite{PoseTrack} is a dataset consisting of video sequences with multiple people. The dataset aims to estimate the pose and track humans in video challenges. The annotations of the video frames contain 17 human body keypoints per person, similar to the original COCO dataset. 

In addition to these datasets, there are other human-oriented datasets, e.g., the DAVIS dataset \cite{perazzi2016} with people in challenging poses or the EPIC-KITCHENS dataset \cite{Damen2021PAMI} with egocentric views. However, they lack human keypoint annotations. The overview of the datasets is shown in \tabref{tab:dataset_summary}.

\begin{table}[h]
\centering
\setlength\tabcolsep{5pt}
    \begin{tabular}{l||ccc|c|r} 
    \multirow{2}{*}{\textbf{Dataset}} & \multicolumn{3}{c|}{Keypoints} & \multirow{2}{*}{Data type} & Size\\
    & body & hand & face  &  & [frames] \\ \hline
    COCO & 17 & - & - & image & 200 000\\ \hline
    COCO-WholeBody & 23 & 42 & 68 & image &  66 700\\ \hline
    Halpe & 26 & 42 & 68 & image & 43 700\\ \hline
    MPII & 16 & - & - & image & 25 000\\ \hline
    DensePose & 17 & - & - & video & 50 000\\ \hline
    DAVIS & - & - & - & video & 3 455\\ \hline
    EPIC-KITCHENS & - & - & - & video & 20 000 000\\ \hline
    \end{tabular}
    \caption{\small Properties of the presented datasets.}
    \label{tab:dataset_summary}
    \vspace*{-7mm}
\end{table}


\subsection{Human keypoint detectors}

\subsubsection{Keypoint detection from RGB images}
In general, there are two approaches for the detection of human body keypoints in 2D images. 
\paragraph{\textbf{Bottom-up methods}} detect the human body keypoints at first and they are reconstructed (connected) to each human body pose (skeleton) afterwards. 
The main representative of this approach is OpenPose \cite{openpose, simon2017hand}. Specific models of MMPose~\cite{mmpose2020} also use the bottom-up approach (not used here). The bottom-up approach is faster than the top-down method with more people appearing in the input images, as the human keypoints are detected all at once.
\paragraph{\textbf{Top-down methods}} proceed in the reverse order, as they first detect people and their bounding boxes and the desired body keypoints are detected separately in each of the detected human body bounding box. 
Such approach is used in AlphaPose \cite{fang2017rmpe, li2018crowdpose, xiu2018poseflow}, MMPose~\cite{WangSCJDZLMTWLX19}, and Detectron2~\cite{wu2019detectron2} / DensePose~\cite{guler2018densepose}. 

There are also specialized detectors for specific body parts. Here we employ  MediaPipe~\cite{mediapipe} for detection of keypoints on the hands.
In this work, we compare the following pose estimators for human keypoint detection in the close proximity scenario (the overview is shown in \tabref{tab:detectors_summary}):
\begin{itemize}
    \item OpenPose \cite{openpose, simon2017hand} -- Real-time multi-person system to detect human keypoints. It is divided into 3 detection blocks (body+foot, hand, face). The first block is trained on COCO and MPII \cite{andriluka14cvpr} datasets. The hand and face blocks are trained on the MPII dataset. The latest version supports single-person tracking.  
    \item Detectron2 \cite{wu2019detectron2} -- A library that provides solutions to different computer vision tasks, including human keypoint detection (originally DensePose \cite{guler2018densepose}). The detector was trained on the original COCO dataset and detects only \texttt{body} keypoints. Another drawback is the lack of a tracking feature. 
    \item MMPose - model HRNetV2-W48 \cite{WangSCJDZLMTWLX19} -- A selected model for the estimation of human pose from the OpenMMLab project. It contains models for the annotation of all three groups (\texttt{body}, \texttt{hand}, \texttt{face} keypoints). The models are trained on the original COCO dataset and then fine-tuned using the COCO-WholeBody dataset. It also provides tracking of poses in the images. 
    \item AlphaPose \cite{fang2017rmpe, li2018crowdpose, xiu2018poseflow} -- Multi-person pose estimator and tracker with models trained either on the COCO-WholeBody dataset or on the Halpe dataset.
    \item MediaPipe -- Open source framework introduced in \cite{mediapipe}, containing a hand keypoint detection called MediaPipe Hands \cite{mediapipe_hands}. The hand detector provides output similar to the Halpe and COCO-WholeBody ones.
\end{itemize}

\begin{table}[tb]
\centering
    \begin{tabular}{l||rrr|c|c}
    \multirow{2}{*}{\textbf{Detector}} & \multicolumn{3}{c|}{Keypoints for 1 person} & Person & Training\\
    & body & hand & face  & tracking & dataset \\ \hline
    OpenPose & 25 & 42 & 68 & single & COCO + MPII\\ \hline
    Detectron2 & 17 & - & - & -  & COCO\\ \hline
    MMPose & 23 & 42 & 68 & multi & COCO + Halpe\\ \hline
    AlphaPose & 26 & 42 & 68 & multi & Halpe\\ \hline
    MediaPipe & - & 42 & - & multi & private\\ \hline
    \end{tabular}
    \caption{\small Properties of tested human keypoint detectors.}
    \label{tab:detectors_summary}
 \vspace*{-6mm}
\end{table}

\subsubsection{From keypoints in 2D to 3D}
For several applications such as HRI, keypoint coordinates in 3D rather than in the image are necessary. Typically, pixel coordinates from 2D keypoint detectors are complemented by their depth from RGB-D sensors 
or stereo cameras. 
The approaches in \cite{rim2020real, zimmermann20183d, pascual20223dhuman} predict 3D human body keypoints based on 2D detections and depth information. Fang et al.~\cite{fang2021_3dhuman} combine 2D human body keypoints detection in the RGB images with fitting a 3D human body model to obtain the resulting 3D keypoints.   
Nguyen et al.~\cite{nguyen2018} present a solution detecting the 2D human body keypoints and transforming them directly to the 3D positions based on a disparity map constructed from two cameras.

\subsubsection{Human 3D body surface reconstruction}
As an alternative to depth information from RGB-D or stereo, complete reconstruction of human body surface in 3D with 2D keypoints as input has been developed in the Skinned Multi-Person Linear model \cite{loper2015smpl, bogo2016keep, kanazawa2018end}, improved in SMPL-X~\cite{pavlakos2019expressive}. 
Additional works in \cite{omran2018neural, pavlakos2018learning} build on top of the 3D \textit{SMPL} human body model for the human surface detection from a single RGB image. In \cite{omran2018neural},  the human body is divided into 12 semantic parts, which an encoding CNN processes to directly predict 3D SMPL body model parameters.  
An alternative approach, not using SMPL or its derivatives, is DensePose \cite{guler2018densepose}.

A drawback of these approaches for HRI is that the scale of the 3D reconstructed human body model is not known.




This work is based on a student thesis \cite{docekal_thesis}, in which additional details of specific aspects can be found, but contains new results and analyses.

\section{Materials and methods}
\label{sec:met}
\subsection{New close proximity human keypoint dataset}
\label{subsec:dataset}
None of the mentioned datasets contains people captured in close proximity. Therefore, we created a HiCP dataset with wholebody keypoint annotations, which is publicly available \url{https://osf.io/qfkvt/} \cite{dataset}. 
The approach to generating the dataset was to crop anotated bodies visible in images that are both in the COCO and Halpe datasets, to avoid any detector bias. The annotations are from the Halpe dataset, as they contain 136 keypoints compared to COCO's 133. This approach is also applicable to other datasets.

\begin{figure}[tb]
    \centering
    \includegraphics[width=0.38\textwidth]{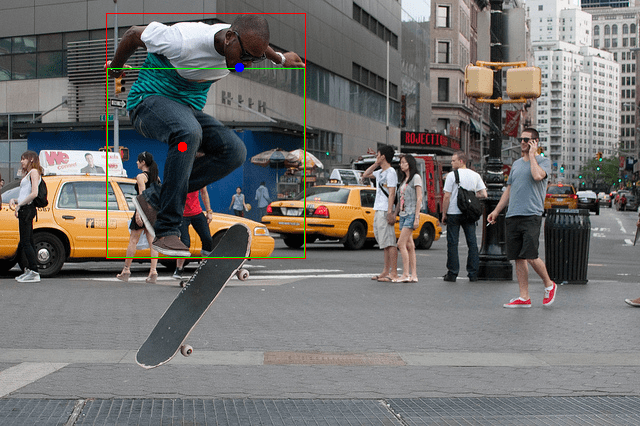}
    \caption{\small
    HiCP dataset images are crops from Halpe images.
    The crops are marked by a bounding box (red, human head; green, body only). The bounding boxes are found automatically as a function of  the position of the head (blue circle) and the  body (red circle). 
    }
    \label{fig:custom_dataset}
    \vspace*{-6mm}
\end{figure}

The creation procedure starts with cropping out all annotated persons from an image based on their annotated bounding boxes. 
Then, the keypoint annotations are also moved accordingly to fit the cropped image correctly. The images created in this way form a subset of the dataset called \textit{Basic}. Then, we remove each person's head to create another subset of the dataset, called \textit{Headless}. Finding the head is straightforward as the keypoints of the head are usually provided. The 2D position of the rest of the human body with respect to the head is determined as a mean of the rest of the \texttt{body} keypoints present in the annotation. The proposed method is shown in \figref{fig:custom_dataset}.

The main disadvantage of this method is the resulting resolution downscale. Original images often contain multiple people in the background, leading to a very low resolution for each cropped image. Therefore, we chose only cropped images with at least 20,000 pixels. The \textit{Basic} and \textit{Headless} subsets contain 1624 and 1608 images, respectively, with corresponding keypoint annotations, and are available at \cite{dataset}. 

\subsection{2D to 3D conversion}
\label{subsec:conv}
For our experiments, we used a robotic platform consisting of an iCub humanoid robot \cite{icub} and the Intel RealSense D435 camera that provides RGB-D images. The detected human keypoints in RGB images are represented by a vector ($u$, $v$, $c$), where ($u$, $v$) are pixel coordinates and $c$ is a confidence score representing the likeliness that the keypoint exists in the calculated position. Low confidence scores suggest false positive detections. 

The 2D positions of the selected human keypoints are converted to 3D by adding depth information as 
\begin{equation}
    x = \frac{k\left(u - c_x\right)}{f_x}, \qquad 
    y = \frac{k \left(v-c_y \right)}{f_y}, \qquad
    z = k,
\end{equation}
where $k$ is the measured depth and the focal lengths $f_x$, $f_y$ and the coordinates of the principal point $c_x$, $c_y$ are the intrinsic parameters of the camera. 

Instead of computing the 3D position of the keypoint from just the 2D position of the detection, we take into account all points in the neighborhood of detections to compute the final 3D position. The reason behind this approach is to overcome inaccuracies in keypoint detection and depth measurement. This method is similar to a baseline method in \cite{zimmermann20183d} called Naive Lifting.
The size of the neighborhood in the 3D coordinate system is different for each keypoint group---20 mm for \texttt{body} keypoints and 3 mm for \texttt{face} and \texttt{hand} keypoints. The desired neighborhood in 2D is determined by the projection as 
\begin{gather}
    x_1 = \frac{k \left( u_1 - c_x \right)}{f_x}, \qquad
    x_2 = \frac{k \left( u_2 - c_x \right)}{f_x}, \\
    x_1 - x_2 = \frac{k \left( u_1 - c_x \right)}{f_x} - \frac{k \left( u_2 - c_x \right)}{f_x}, \\
    \frac{f_x}{k}(x_1 - x_2) = u_1 - u_2.
\end{gather}

The expression $(x_1 - x_2)$ corresponds to the desired distance in the 3D space and the expression $(u_1 - u_2)$ is the resulting distance in the 2D image plane on the $x$-axis. The distance in the $y$-axis is done in the same way.

The pixels in the derived neighborhood are converted to the 3D positions in the camera frame. The 3D positions of the detected keypoints in the camera frame are the median coordinates of the 3D positions in the neighborhood.
As a final step, the 3D keypoints are transformed from the camera coordinate frame to the robot base coordinate frame.

\subsection{Evaluation of 2D detections}
\label{subsec:eval}
We chose Object Keypoint Similarity (OKS) \cite{oks} as a metric for the evaluation of 2D keypoint detection. It is a similar metric to the Intersection over Union (IoU) used for object detection evaluation.
OKS is computed from the detected keypoints and their ground-truth annotations based on their distance and object scale as
\begin{equation}
    \mathrm{OKS} = \frac{\sum_{i} \exp{\left (\frac{-d_i^2}{2s^2k_i^2}\right )} \delta(v_i > 0)}{\sum_{i} \delta (v_i > 0)},
\end{equation}
where $d_i$ is the distance between the detected keypoint and its corresponding ground truth position, $s$ is an object (person) scale, and $k_i$ is a keypoint constant to control the falloff. The $\delta$ function is equal to one if the keypoint occurs in the image (i.e., $v_i > 0$, where $v_i$ is the so-called visibility flag of the keypoint) and zero otherwise. 
All values correspond to the keypoint with index $i$ and are summed over all ground-truth annotated keypoints.
The OKS metric shows only how close are the predicted keypoints to the annotated ones (value from 0 to 1).
Therefore, \textit{OKS} threshold is used to convert the detection evaluation to a classification problem. If the calculated OKS is higher than the \textit{OKS} threshold, it is considered a true positive detection; otherwise, it is considered a false positive detection.
We use the same \textit{OKS} thresholds as in the COCO challenge, which are $0.5$, $0.75$, and the interval from $0.5$ to $0.95$ with a step value $0.05$. This means that we sample the interval with the mentioned step value and compute precision and recall for all these \textit{OKS} threshold values. Their means are the Average Precision (AP) and Average Recall (AR) over this interval.



\subsection{Demonstrator -- real-time whole-body avoidance on a humanoid robot}
To demonstrate the framework developed here in action, we deployed the different variants of the human pose detection pipeline in a closed-loop scenario on the iCub humanoid robot. We draw on the work of Roncone et al.~\cite{Roncone_IROS_2015,Roncone2016} whereby the robot has a distributed representation of its peripersonal space. Every keypoint of the human detected by the camera is transformed into 3D coordinates, remapped into the robot base frame, and finally into the reference frame of every module of electronic skin covering the complete robot body. The distance from the robot skin surface gives rise to activation patters, from which the centroids---the most ``threatened'' locations on each body part of the robot---are calculated (the distance only \cite{Roncone_IROS_2015}, not the distance and time to contact \cite{Roncone2016}, variant is used here). These ``Cartesian obstacles'' are then remapped into the joint space and fed into a whole-body reactive controller similar to the one used in \cite{nguyen2018}. 


\section{Results}
Our results consist of an analysis of the performance of four detectors---AlphaPose, OpenPose, MMPose, and the combination of Detectron2 (only \texttt{body} keypoints) and MediaPipe (only \texttt{hand} keypoints). First, we evaluate the 2D detection precision on our HiCP dataset (Sec.~\ref{subsec:dataset}). Second, we evaluate the accuracy of the 3D positions of the keypoints in two different experiments. Then, we analyze their failures. Finally, we demonstrate the proposed solution in a human-robot interaction scenario. 

\subsection{Keypoints detection}
In this case, we have ground-truth data from the annotated HiCP dataset; thus, we can evaluate the differences between the detected and ground-truth positions. 
The detections are evaluated using the average precision (AP) and average recall (AR) metric using Object Keypoint Similarity (OKS). AP and AR are computed separately for the \texttt{body} and \texttt{hand} keypoints. 

\begin{figure}[tb]
    \centering
    \begin{subfigure}[t]{.23\textwidth}
    \centering
    \includegraphics[width=\textwidth]{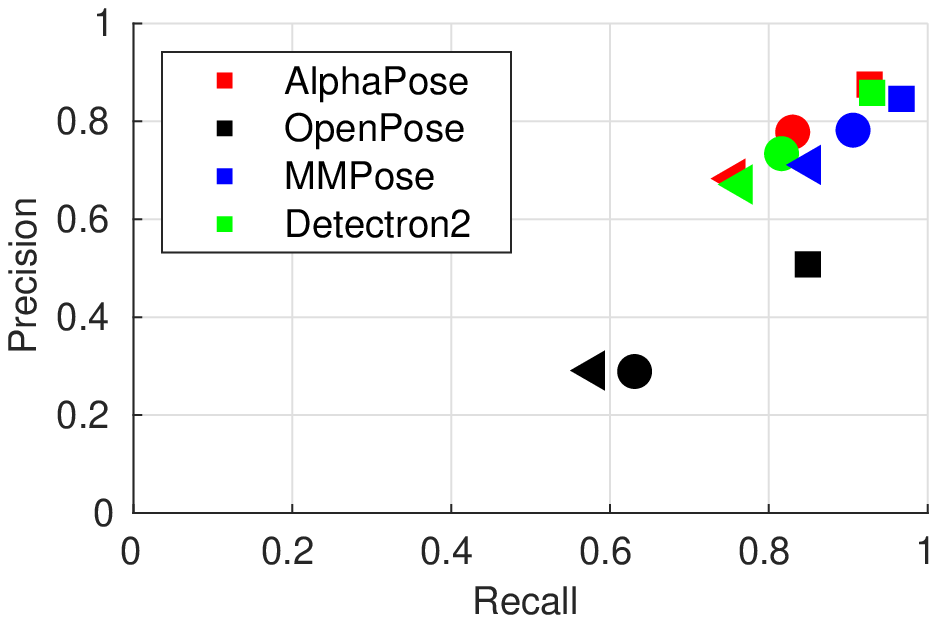}
    \caption{\small \texttt{Body} keypoints on \textit{Basic} subset of the HiCP dataset.}
    \end{subfigure} \hfill %
    \begin{subfigure}[t]{.23\textwidth}
    \centering
    \includegraphics[width=\textwidth]{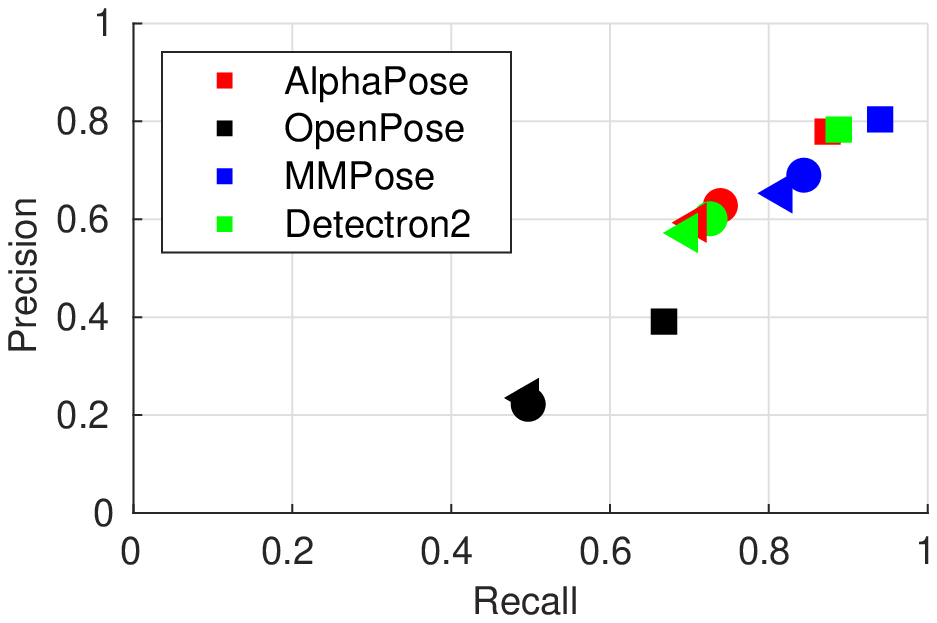}
    \caption{\small \texttt{Body} keypoints on \textit{Headless} subset of the HiCP dataset.}
    \end{subfigure}
    \begin{subfigure}[t]{.23\textwidth}
    \centering
    \includegraphics[width=\textwidth]{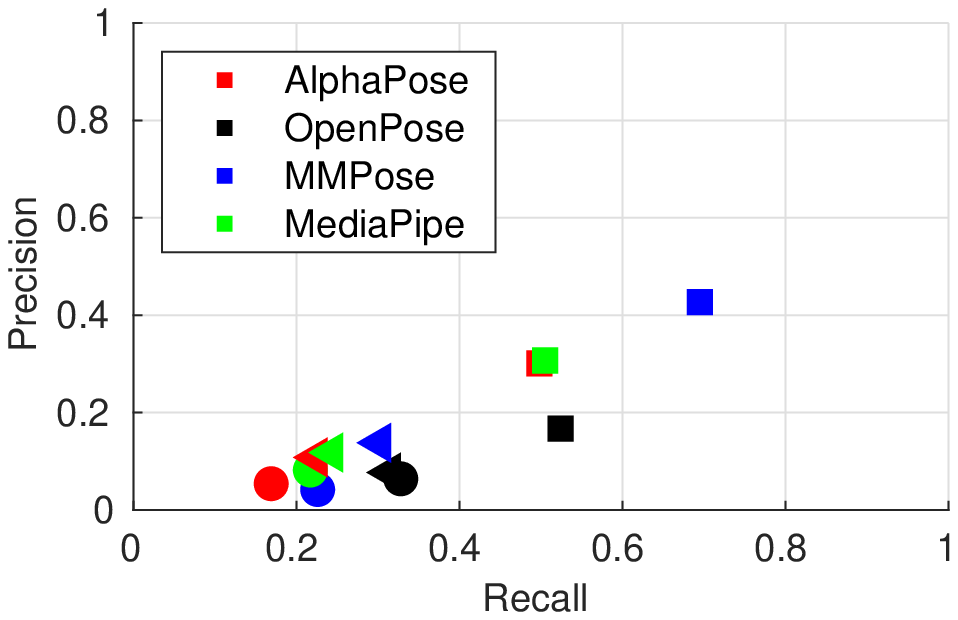}
    \caption{\small \texttt{Hand} keypoints on \textit{Basic} subset of the HiCP dataset.}
    \end{subfigure}\hfill 
    \begin{subfigure}[t]{.23\textwidth}
    \centering
    \includegraphics[width=\textwidth]{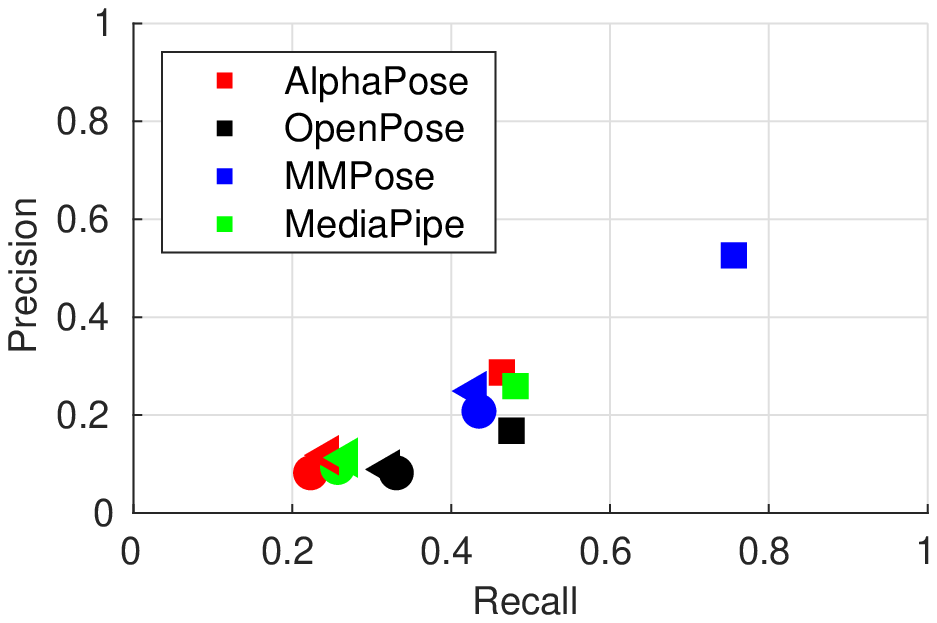}
    \caption{\small \texttt{Hand} keypoints on \textit{Headless} subset of the HiCP dataset.}
    \end{subfigure}%
    \caption{\small Precision-Recall of 2D detections, all detectors. Symbols encode the spatial accuracy \textit{OKS} threshold---0.5 (square), 0.75 (circle) and the interval from $0.5$ to $0.95$ with a step value $0.05$ (triangle). For details on {\it OKS}, see Sec.~\ref{subsec:eval}. }
    \label{fig:2d_comp}
     \vspace*{-3mm}
\end{figure}

Precision-Recall scatter plot for \texttt{body} and \texttt{hand} keypoints and both datasets is in \figref{fig:2d_comp}. 
The OpenPose detector is outperformed by other detectors; only the recall values for \texttt{hand} keypoint detection are better than those for AlphaPose and MediaPipe, but the precision is worse.\footnote{OpenPose overall detects more \texttt{hand} keypoints than other detectors.} AlphaPose and Detectron2 perform similarly in \texttt{body} keypoint detection; AlphaPose and MediaPipe in \texttt{hand} keypoint detection. 
MMPose detector has higher recall than the other detectors. In case of hand detection with 0.5 \textit{OKS} threshold, the difference is significant. Nevertheless, the precision of MMPose detections is higher only for the \textit{Headless} dataset, suggesting that its performance is less degraded by the absence of the head in the image. 
In summary, the MMPose detector has the best results and OpenPose the worst results. 

\begin{figure}[tb]
    \centering
    \begin{subfigure}[t]{.12\textwidth}
    \centering
    \includegraphics[width=\textwidth]{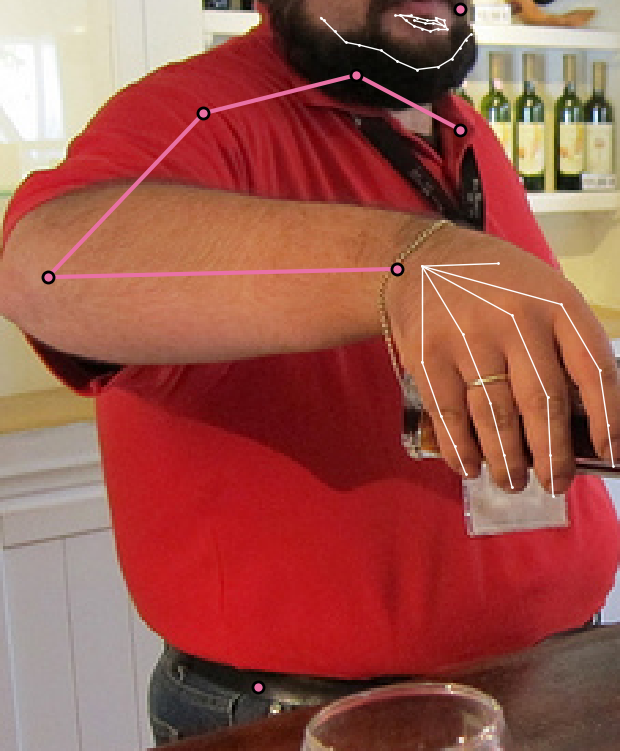}
    \caption{\small AlphaPose.}
    \label{fig:example_alphapose}
    \end{subfigure}%
    \begin{subfigure}[t]{.12\textwidth}
    \centering
    \includegraphics[width=\textwidth]{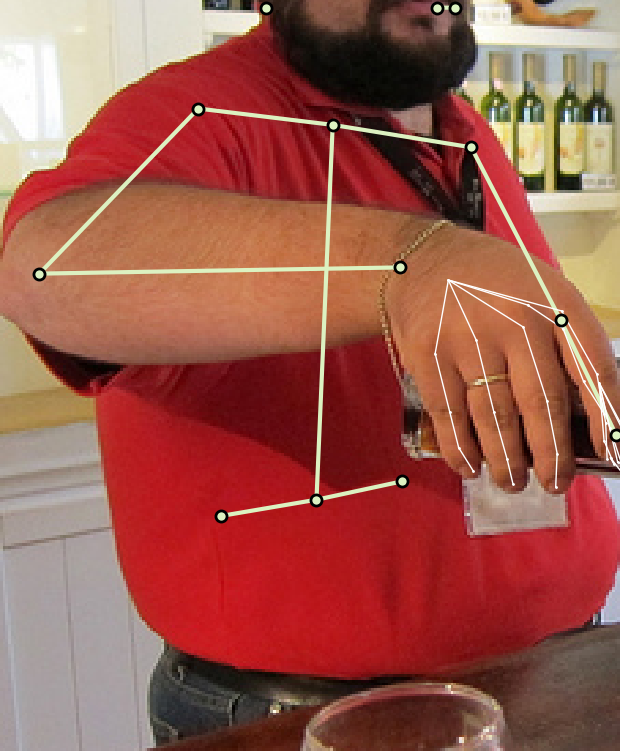}
    \caption{\small OpenPose.}
    \label{fig:example_openpose}
    \end{subfigure}%
    \begin{subfigure}[t]{.12\textwidth}
    \centering
    \includegraphics[width=\textwidth]{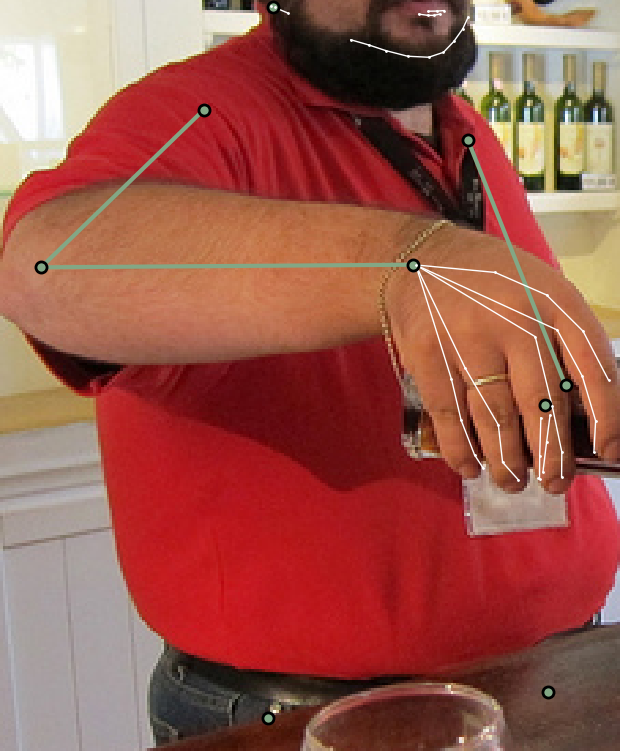}
    \caption{\small MMPose.}
    \label{fig:example_mmpose}
    \end{subfigure}%
    \begin{subfigure}[t]{.12\textwidth}
    \centering
    \includegraphics[width=\textwidth]{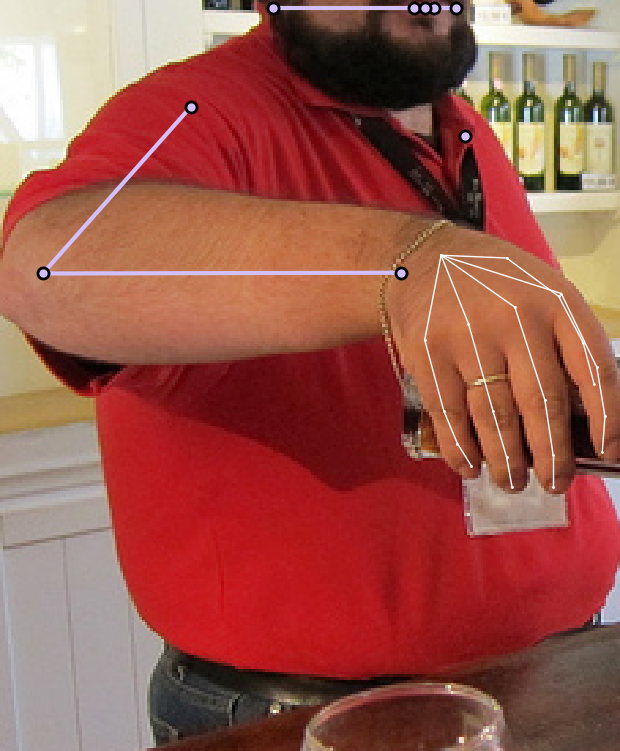}
    \caption{\small Detectron2 + MediaPipe.}
    \label{fig:example_detectron2_mediapipe}
    \end{subfigure}
    \vspace*{-2mm}
    \caption{\small Detected human keypoints on an image from the \textit{Headless} subset (head not visible),  of the HiCP Dataset.}
    \label{fig:example_annotation}
    \vspace*{-6mm}
\end{figure}

Figure \ref{fig:example_annotation} shows example annotations from all detectors on an image from the \textit{Headless} HiCP dataset. 
In this visual comparison, OpenPose detections are the worst, as hip keypoints are detected in the abdomen in \figref{fig:example_openpose}. Although the MMPose detector has the best detection results, its finger keypoint detection is mediocre---in \figref{fig:example_mmpose}, the fingers look like they are broken and often even outside the correct position of the finger. This may cause problems in 3D position computation due to incorrect depth information for such keypoints.

\subsection{Relative 3D position evaluation}
\label{subsec:res_rel}
We prepared an experiment for 3D position evaluation where a human is in a static pose with respect to the robot base coordinate frame. The robot moves with its torso and head to observe the human from 100 predefined positions. The desired output is that the human keypoints have the same 3D positions with respect to the robot base coordinate frame during the whole experiment.
We calculated the distances of the detected keypoints from their center position obtained as the medians of individual coordinates. 
This approach is able to overcome the lack of ground-truth 3D positions, as we compare just the deviations and not the exact 3D positions.

\begin{figure*}[tb]
    \centering
    \includegraphics[width=0.235\textwidth]{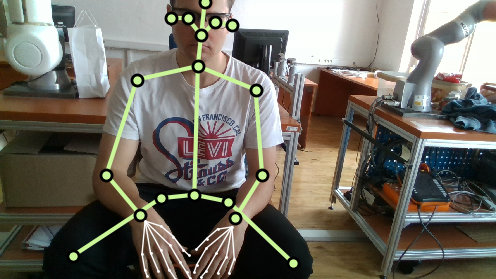}\hfil
    \includegraphics[width=0.72\textwidth]{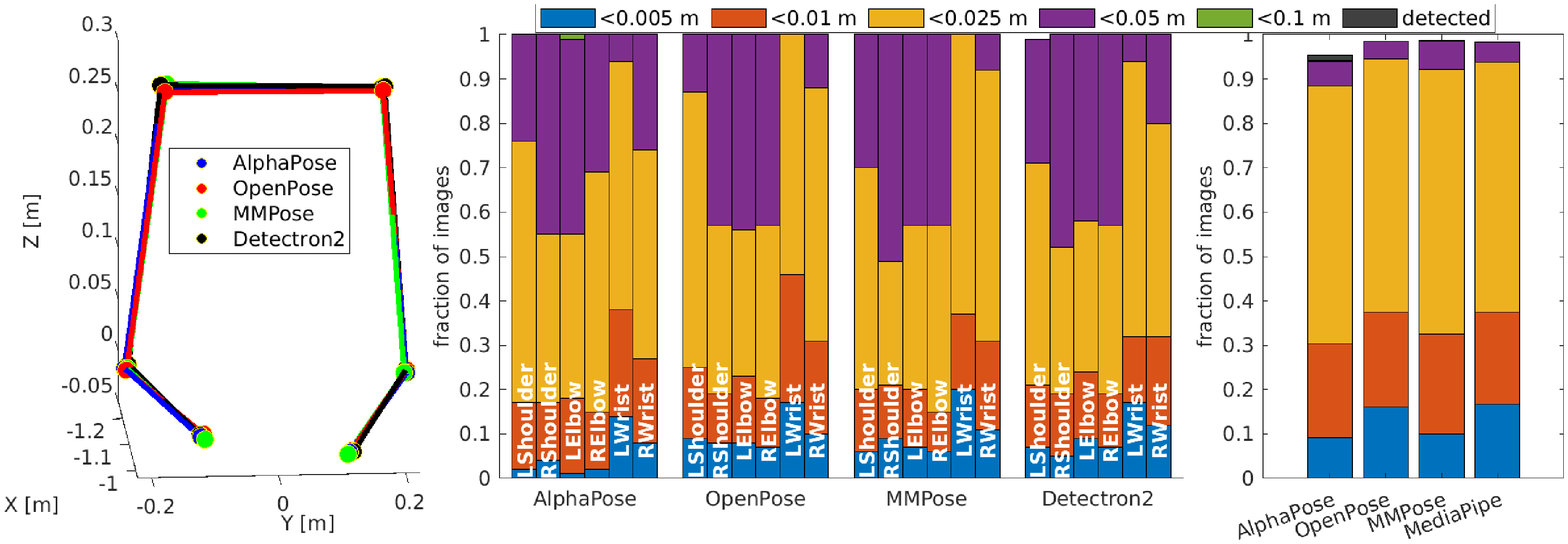}
    \includegraphics[width=0.235\textwidth]{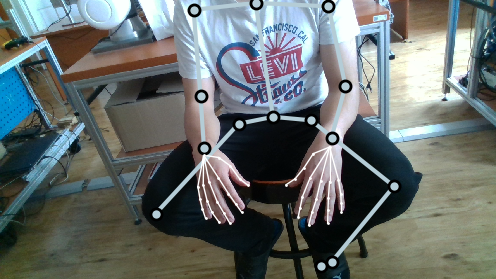}\hfil
    \includegraphics[width=0.72\textwidth]{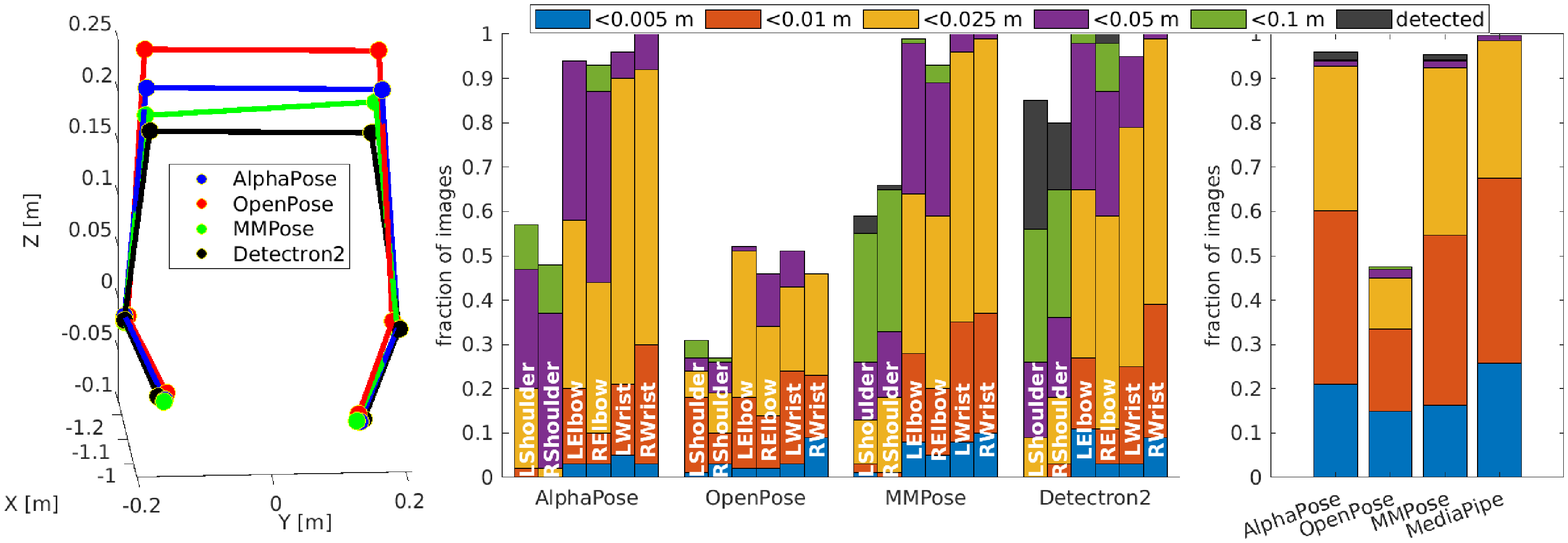}
    \caption{\small The upper-view (top) and lower-view (bottom) experiments. From left to right: an image with AlphaPose detections; median keypoint positions skeletons; the fraction of images where upper body keypoints were detected with confidence $>$ 0.1 within the specified color-coded distance from their median positions
    ; the fraction of images (mean over the keypoints) where the \texttt{hand} keypoints are detected within the specific distance from their median positions.}
    \label{fig:rel_upper}
    \vspace*{-6mm}
\end{figure*}

The camera was adjusted to two different poses with different field of view. In the first pose (Upper-view experiment), the camera was heading directly forward with no tilt, which resulted in observing mainly the upper body and the head of the participant (see \figref{fig:rel_upper} top). On the other hand, the second pose (Lower-view experiment) focuses on the core and lower part of the human body, as the camera was tilting down (see \figref{fig:rel_upper} bottom).

Figure \@~\ref{fig:rel_upper} shows the camera views for both camera poses, the median positions of the keypoints visualized in the form of skeletons, and the fraction of images where the upper body (shoulders, elbows, wrists)  and \texttt{hand} keypoints were detected with confidence $>$ 0.1 from their median positions.
We cannot directly compare the detectors, as each of them placed the keypoints in different positions. We can compare them only based on the stability of the detections.

In the Upper-view experiment, there are almost no differences between the detectors. The results are more interesting in the Lower-view experiment. OpenPose is the detector with the lowest fraction of detections---the maximum is around 50 \%. The other three detectors have similar distributions of distances from the median position; however, the median positions are very different; e.g., the line connecting the shoulder keypoints detected by MMPose is not horizontal; other detectors have horizontal lines, but there is an offset in its position. In the case of \texttt{hand} keypoints detection, the MediaPipe detector outperforms the other ones, as it is capable of detecting hands in all images.


\subsection{Absolute 3D position evaluation}
\label{subsec:res_abs}
This experiment compares the 3D positions measured by an external system and the estimated 3D positions in the camera frame provided by our pipeline. The external system is the Qualisys Motion Capture (MoCap) System \cite{qualisys}. Special markers are placed on the body keypoints and then localized. 
The MoCap System can observe the moving human during the experiment.
We computed the error (distance) between the detected keypoints in 3D and their correspondences in MoCap data to determine the precision of the detection of 2D human keypoints combined with the transformation to 3D positions. 

\begin{figure*}[tb]
    \centering
    \includegraphics[width=0.3\textwidth]{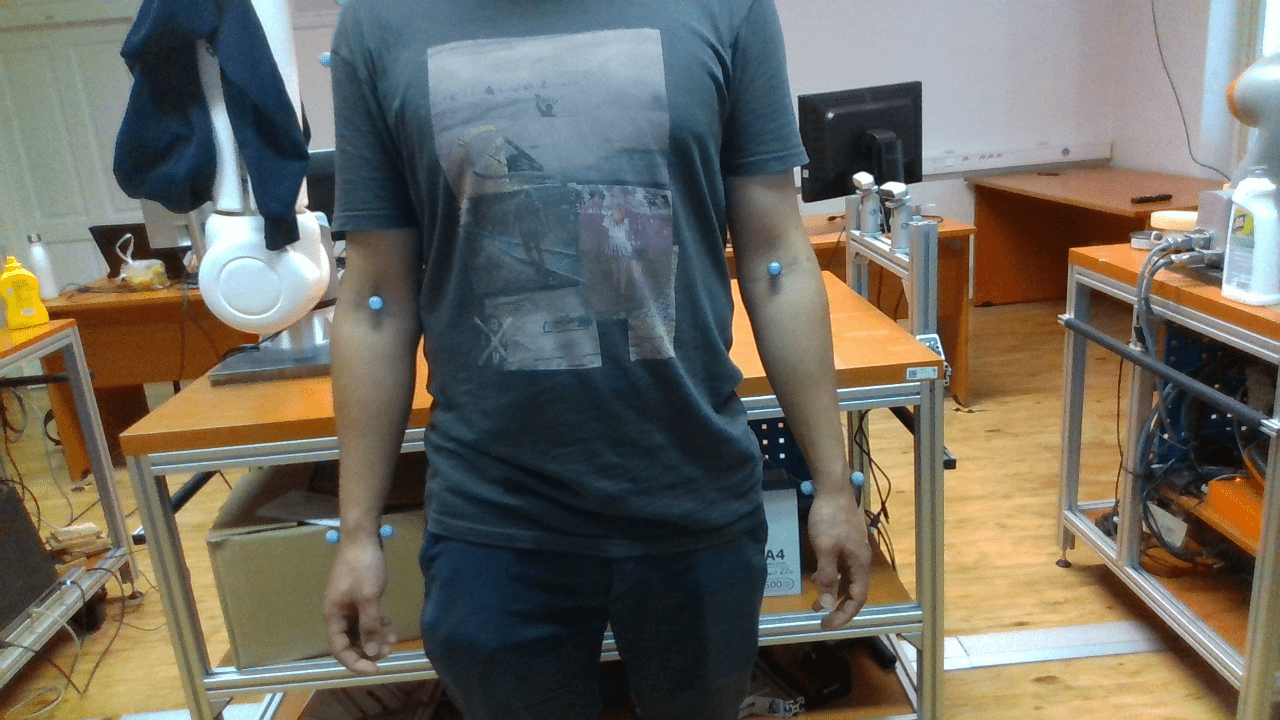}
     \includegraphics[width=0.3\textwidth]{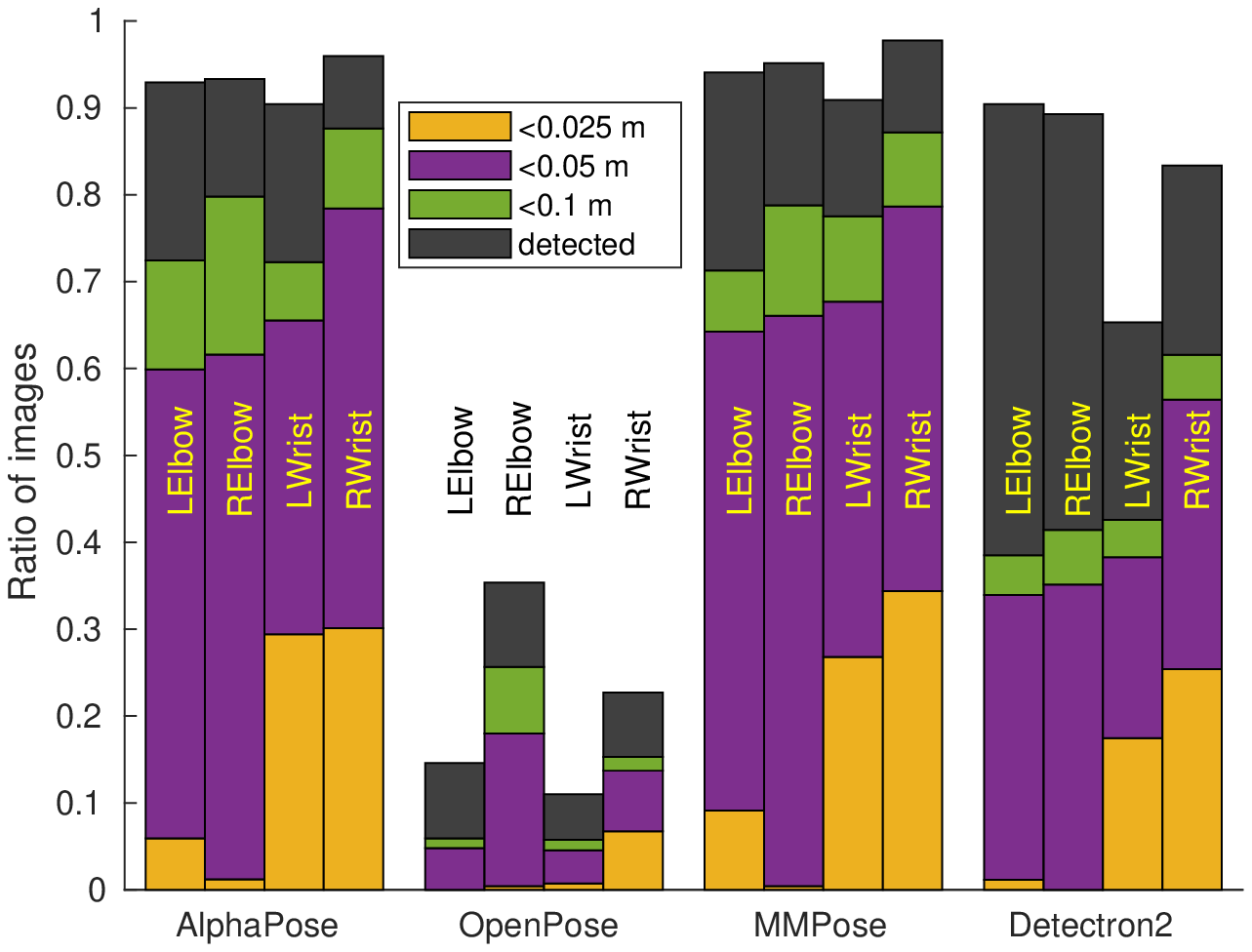}
     \includegraphics[width=0.36\textwidth]{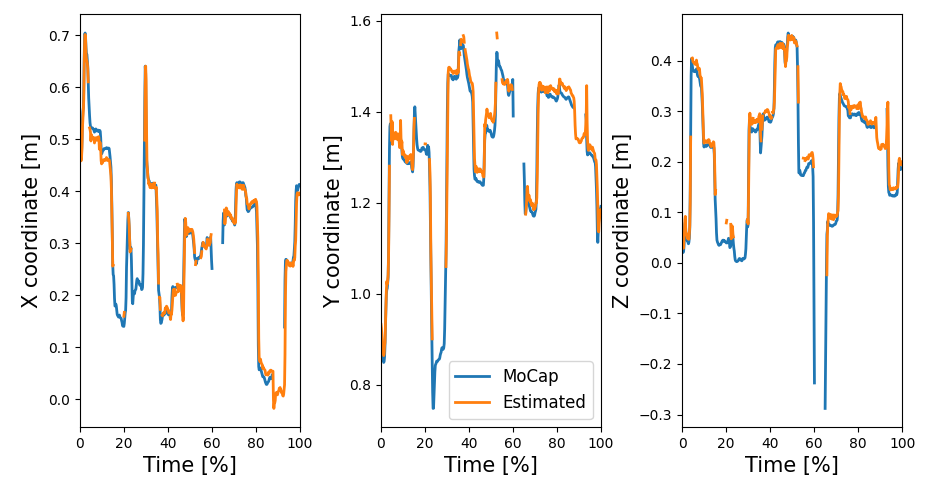}
     \vspace*{-1mm}
    \caption{\small The MoCap Experiment. A human pose with MoCap markers -- small balls on the body (left);
    Fraction of images where upper body keypoints were detected with confidence $>$ 0.1 within the specified distance from their MoCap correspondences (center);
    Position of the right wrist keypoint as estimated by the RGB-D camera (red curve) and the MoCap system (blue curve), on the right.}
    \label{fig:mocap_exp}
    \vspace*{-6mm}
\end{figure*}

We equip the human body with 12 reflective markers at six keypoints: shoulders, elbows, and wrists (see \figref{fig:mocap_exp} left). Two markers are placed on every keypoint to ensure visibility by the MoCap system. The ground truth position of a keypoint is computed as the mean of both positions detected by the MoCap. An additional marker is placed on the Intel RealSense camera to obtain the estimation of translation between the camera frame and the MoCap reference frame.
To calculate the transformation between the camera and MoCap frame, we manually annotate eight markers in two images from the Intel RealSense camera and compute the transformation using an algorithm introduced by Arun et al.~\cite{arun87}. The translation part of the transformation corresponds to the position of the marker located on the RealSense camera.

Figure \ref{fig:mocap_exp} shows one of the human poses during the experiment, the fraction of images where upper body keypoints were detected with confidence $>$ 0.1 within a specific distance (0.025, 0.05, and 0.1 m) from their MoCap correspondences and the 3D position of the right wrist keypoint. The number of successful detections and the median distances are shown in Tabs.\@~\ref{tab:mae_alpha}, and \ref{tab:mae_alpha01}. 

\begin{table}[tb]
\centering
\setlength\tabcolsep{5pt}
\begin{tabular}{l||rr|rr|rr|rr} 
\multirow{2}{*}{\textbf{Keypoint}}& \multicolumn{2}{c|}{\textbf{AlphaPose}}  & \multicolumn{2}{c|}{\textbf{MMPose}} & \multicolumn{2}{c|}{\textbf{Detectron2}} & \multicolumn{2}{c}{\textbf{OpenPose}}\\ \cline{2-9}
& \# & Med. & \# & Med. & \# & Med. & \# & Med. \\ 
& [-] & [cm] & [-] & [cm] & [-] & [cm] & [-] & [cm] \\ \hline
     LElbow & 443 & 4.0 & 419 & 3.4 & 149 & 12.0 & 29 & 11.3\\ \hline
     RElbow & 424 & 4.2 & 456 & 4.2 & 175 & 11.2 & 134 & 4.8\\ \hline
     LWrist & 385 & 2.7 & 408 & 2.8 & 133 & 3.1 & 20 & 3.4\\ \hline
     RWrist & 424 & 2.7 & 430 & 2.7 & 146 & 3.1 & 55 & 2.5\\ \hline
\end{tabular}
\caption{\small Keypoint detection: the number of points with confidence above 0.3  (\#), and the median distances (Med.) to their MoCap ground truth location. 
}
\label{tab:mae_alpha}
\vspace*{-4mm}
\end{table}

\begin{table}[tb]
\centering
\setlength\tabcolsep{5pt}
\begin{tabular}{l||rr|rr|rr|rr} 
\multirow{3}{*}{\textbf{Keypoint}}& \multicolumn{2}{c|}{\textbf{AlphaPose}} & \multicolumn{2}{c|}{\textbf{MMPose}} & \multicolumn{2}{c|}{\textbf{Detectron2}} & \multicolumn{2}{c}{\textbf{OpenPose}} \\ \cline{2-9}
& \# & Med. & \# & Med. & \# & Med. & \# & Med. \\
& [-] & [cm] & [-] & [cm] & [-] & [cm] & [-] & [cm] \\ \hline

     LElbow & 464 & 4.0 & 468 & 3.5 & 440 & 16.7 & 66 & 15.4\\ \hline
     RElbow & 462 & 4.3 & 471 & 4.2 & 442 & 11.5 & 175 & 5.0\\ \hline
     LWrist & 454 & 2.9 & 457 & 2.9 & 332 & 3.3 & 48 & 5.6\\ \hline
     RWrist & 469 & 2.8 & 468 & 2.8 & 399 & 2.9 & 119 & 3.0\\ \hline
\end{tabular}
\caption{\small Keypoint detection: the number of points with confidence above 0.1  (\#), and the median distances (Med.) to their MoCap ground truth location. }
\label{tab:mae_alpha01}
\vspace*{-6mm}
\end{table}

Like in previous experiments, OpenPose performs the worst since it detects at least one of the compared keypoints in only 40 \% of the images. Interestingly, OpenPose detected the right arm keypoints better than the left arm ones. Most of the Detectron2 detections have confidence lower than 0.3, and they are far away from the MoCap correspondences; e.g., more than half of the detections of the elbows are farther than 0.1 m from their correspondences, and the median distance is around 12 cm. These high distances may cause low keypoint confidence scores. AlphaPose and MMPose are very similar, as the number of detections, the distance distribution, and the median distance are almost the same and remarkably outperform the other two detectors. 

The highest errors are for shoulder detection, as it is the largest joint, and the marker points were placed on the outer part of the shoulder joint, whereas the detection usually detects the shoulder in the inner part of the joint. Therefore, we omit them from the comparison. 

The positions of the right wrist keypoint are plotted in \figref{fig:mocap_exp} (right). The discontinuities are caused by either not detecting the keypoint in the RGB image from the RealSense camera as they did not occur in the images or the marker points were not visible in the MoCap system. 

\subsection{Detectors failure analysis}
As human pose detectors are typically trained on images where human figures are completely visible, we were interested in the limitations of close proximity detection. For example, the OpenPose detector, the only bottom-up detector in our comparison, is sensitive to the visibility of body parts. On the other hand, the rest of the tested detectors use the top-down detection method; thus, they do not depend on the visibility of body parts.

At first, we analyzed the detectors on images from the HiCP dataset, where only parts of the human body are visible. OpenPose detector was able to annotate those where the whole upper body is visible. Except for that, several other images (only head, no left arm, no right arm, or no head) were annotated too, but the AP and AR are much lower than for the whole upper body. Therefore, its performance is inferior to the other detectors using the top-down approach. 

AP and AR are the highest for all other detectors in images with visible whole upper bodies. MMPose has the best values, followed by Detectron2 and AlphaPose. In other cases (only legs, no head, only left arm, only both arms, both arms without shoulders), the results are worse, but MMPose still has the best results, followed by AlphaPose and Detectron2. 
There are only two exceptions where Detectron2 outperformed MMPose---only head and only right arm. 


When we analyze the results from our 3D experiments, there are several findings on the failure of the tested detectors. MMPose detector often tries to add keypoints even if the specific joint is on the edge of the image or not visible. Nevertheless, the confidence score of these keypoints is low. The biggest drawback of this detector is the hand pose. The hand is easily detected, but the detected fingers are not in a natural pose. 
AlphaPose detections seem to be the most conservative---not trying to detect border keypoints, resulting in fewer detections than by MMPose. Moreover, the hand poses look natural.
Detectron2 detector struggles with shoulder detection, even when the shoulders are perfectly visible (Upper-view experiment)---the confidence scores of shoulder keypoints are low. 
OpenPose detections are not stable. For example, different keypoints are detected in a sequence of consecutive images with almost the same view in Lower-view experiment (without tracking). Hand pose detection by OpenPose seems natural; however, the fingers are detected only when the wrist keypoint is detected. 


\subsection{HRI scenario}
\begin{figure}[tb]
    \centering
    \begin{subfigure}[t]{.24\textwidth}
    \centering
    \includegraphics[width=.8\textwidth]{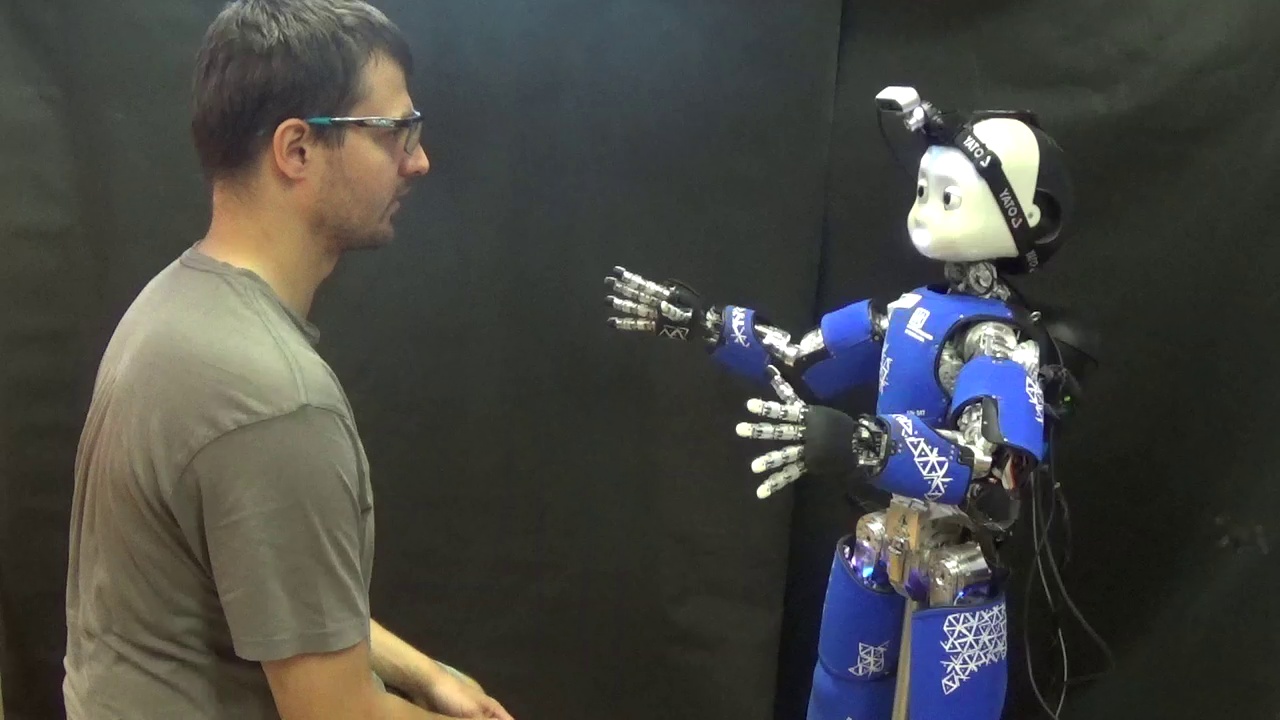}
    \vspace*{-1mm}
    \caption{}
    \label{fig:hold_position}
    \end{subfigure}%
    \begin{subfigure}[t]{0.24\textwidth}
    \centering
    \includegraphics[width=.8\textwidth]{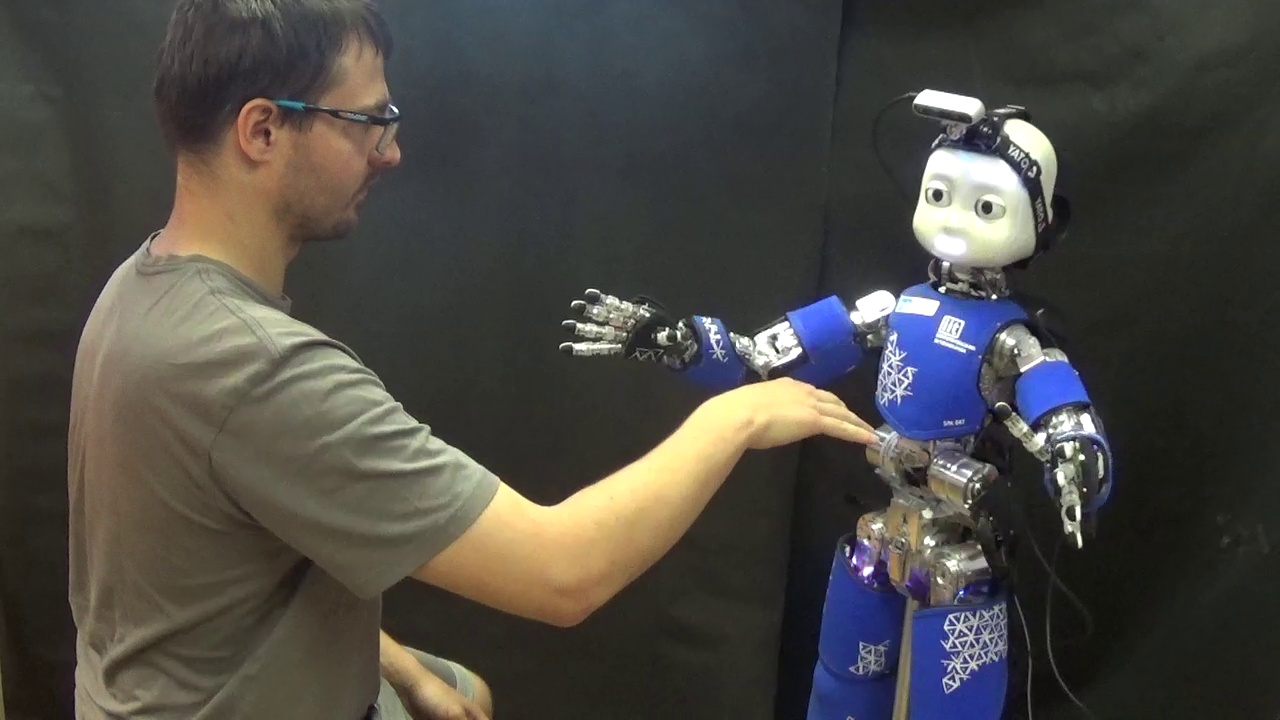}
    \vspace*{-1mm}
    \caption{} 
    \label{fig:avoid_position}
    \end{subfigure}
    \vspace*{-3mm}
    \caption{\small Demonstration of the human body avoidance scenario. Robot in a fixed position (a), avoiding a human hand (b).}
    \label{fig:demonstration}
    \vspace*{-6mm}
\end{figure}

To demonstrate the proposed solution, we prepare a human-robot interaction scenario, where the iCub robot holds the defined position of the left arm. When the participant tries to reach the position, the robot should avoid him and move away as shown in Fig.~\ref{fig:demonstration}. A video of the demonstration experiment is available at \url{https://youtu.be/aSD7mgz4_sE}.
During the experiment, the participant was instructed to perform 3 different poses: 1) sit in front of the robot in its field of view and reach to the robot hand; 2) same as 1) but standing; 3) sit and approach the robot hand with the head.
The experiment shows that in terms of 3D position computation, the results are accurate enough to localize the detected human keypoints to control robot avoidance. 

\section{Conclusion, Discussion and Future work}
We presented HiCP, a new publicly available dataset created by cropping images from the validation set of COCO and Halpe datasets. We quantitatively and qualitatively compared human whole-body keypoint detection methods---OpenPose, MMPose, AlphaPose, Detectron2, and MediaPipe (for \texttt{hand} keypoints)---on the dataset. 
%
We deployed the detectors on a robot with a head-mounted RGB-D camera and evaluated their performance in 3D human keypoint detection in two different experiments:
(i) human in a static pose, (ii) using a motion capture system as a reference. In both cases, the visibility of the human body parts varied. 

We analysed the failure modes of individual detectors. The best performing whole-body keypoint detector in close proximity was MMPose, but it had difficulty with hand/finger pose detection. The OpenPose detector struggled with finding keypoints when the human body was only partially visible.  
The results suggest that the top-down keypoint detection methods (MMPose, AlphaPose, Detectron2) are more suitable for close proximity detection than bottom-up methods (OpenPose).
The MediaPipe detector, which focuses on hands/fingers, outperforms other detectors in 3D \texttt{hand} keypoint experiments; its results on the HiCP dataset were not the best, probably due to the low resolution of the images after cropping.
Thus, we propose a combination of MMPose or AlphaPose for the \texttt{body} keypoints and MediaPipe for the \texttt{hand} keypoints in a single framework---this combination provides the most accurate and robust detection. 
Finally, we demonstrated the framework in a scenario where a humanoid robot interacting with a person used the detected 3D keypoints for whole-body avoidance maneuvers.

Tracking, i.e. taking advantage of the temporal relationships between images in videos, was not employed in this work, with the exception of the MediaPipe detector in the HRI scenario. Some of the other detectors offer tracking options. In preliminary experiments, enabling tracking did not improve performance and slowed down the processing.   

In the future, we plan to finetune the 2D keypoint detectors on the close proximity dataset. The problem with low image resolution after cropping from the original datasets can be alleviated using super-resolution techniques. Specifically, we plan to train the MMPose detector for better handling of hand poses, for example, using images annotated by MediaPipe. Moreover, we plan to improve our 3D keypoint estimation to be more robust to occlusion and noisy detections.
Finally, regarding applications in safety of HRI, RGB-D cameras like Intel Realsense are currently not safety rated. Once such products appear, one will be able to take advantage of them and improve performance of human-robot collaboration tasks (as shown in \cite{lucci2020combining,svarny2019}).





\bibliographystyle{IEEEtran}
\bibliography{close-proximity}

\begin{thebibliography}{10}
\providecommand{\url}[1]{#1}
\csname url@rmstyle\endcsname
\providecommand{\newblock}{\relax}
\providecommand{\bibinfo}[2]{#2}
\providecommand\BIBentrySTDinterwordspacing{\spaceskip=0pt\relax}
\providecommand\BIBentryALTinterwordstretchfactor{4}
\providecommand\BIBentryALTinterwordspacing{\spaceskip=\fontdimen2\font plus
\BIBentryALTinterwordstretchfactor\fontdimen3\font minus
  \fontdimen4\font\relax}
\providecommand\BIBforeignlanguage[2]{{%
\expandafter\ifx\csname l@#1\endcsname\relax
\typeout{** WARNING: IEEEtran.bst: No hyphenation pattern has been}%
\typeout{** loaded for the language `#1'. Using the pattern for}%
\typeout{** the default language instead.}%
\else
\language=\csname l@#1\endcsname
\fi
#2}}

\bibitem{ISO/TS15066}
{ISO, IEC}, ``{ISO/TS 15066 Robots and robotic devices -- Collaborative
  robots},'' {International Organization for Standardization}, Geneva, CH,
  Standard, 2016.

\bibitem{Svarny_SSR_2018}
P.~Svarny, Z.~Straka, and M.~Hoffmann, ``Toward safe separation distance
  monitoring from {RGB-D} sensors in human-robot interaction,'' in
  \emph{International PhD Conference on Safe and Social Robotics (SSR-2018)},
  2018, pp. 11--14.

\bibitem{svarny2019}
P.~Svarny, M.~Tesar, J.~K. Behrens, and M.~Hoffmann, ``{Safe physical HRI:
  Toward a unified treatment of speed and separation monitoring together with
  power and force limiting},'' in \emph{IEEE/RSJ International Conference on
  Intelligent Robots and Systems}, 2019, pp. 7580--7587.

\bibitem{jin2020whole}
S.~Jin, L.~Xu, J.~Xu, C.~Wang, W.~Liu, C.~Qian, W.~Ouyang, and P.~Luo,
  ``{Whole-Body Human Pose Estimation in the Wild},'' in \emph{Proceedings of
  the European Conference on Computer Vision (ECCV)}, 2020.

\bibitem{coco_dataset}
T.-Y. Lin, M.~Maire, S.~Belongie, J.~Hays, P.~Perona, D.~Ramanan,
  P.~Doll{\'a}r, and C.~L. Zitnick, ``{Microsoft COCO: Common Objects in
  Context},'' in \emph{{European Conference on Computer Vision}}, 2014, pp.
  740--755.

\bibitem{fang2017rmpe}
H.-S. Fang, S.~Xie, Y.-W. Tai, and C.~Lu, ``{RMPE}: {Regional Multi-person Pose
  Estimation},'' in \emph{International Conference on Computer Vision}, 2017.

\bibitem{li2020pastanet}
Y.-L. Li, L.~Xu, X.~Liu, X.~Huang, Y.~Xu, S.~Wang, H.-S. Fang, Z.~Ma, M.~Chen,
  and C.~Lu, ``{PaStaNet: Toward Human Activity Knowledge Engine},'' in
  \emph{Conference on Computer Vision and Pattern Recognition}, 2020.

\bibitem{PoseTrack}
M.~Andriluka, U.~Iqbal, E.~Ensafutdinov, L.~Pishchulin, A.~Milan, J.~Gall, and
  S.~B., ``Pose{T}rack: {A} {Benchmark for Human Pose Estimation and
  Tracking},'' in \emph{Conference on Computer Vision and Pattern Recognition},
  2018.

\bibitem{perazzi2016}
F.~Perazzi, J.~Pont-Tuset, B.~McWilliams, L.~{Van Gool}, M.~Gross, and
  A.~Sorkine-Hornung, ``{A Benchmark Dataset and Evaluation Methodology for
  Video Object Segmentation},'' in \emph{Computer Vision and Pattern
  Recognition}, 2016.

\bibitem{Damen2021PAMI}
D.~Damen, H.~Doughty, G.~M. Farinella, S.~Fidler, A.~Furnari, E.~Kazakos,
  D.~Moltisanti, J.~Munro, T.~Perrett, W.~Price, and M.~Wray, ``The
  {EPIC-KITCHENS} {Dataset: Collection, Challenges and Baselines},'' \emph{IEEE
  Transactions on Pattern Analysis and Machine Intelligence (TPAMI)}, vol.~43,
  no.~11, pp. 4125--4141, 2021.

\bibitem{openpose}
Z.~{Cao}, G.~{Hidalgo Martinez}, T.~{Simon}, S.~{Wei}, and Y.~A. {Sheikh},
  ``{OpenPose: Realtime Multi-Person 2D Pose Estimation using Part Affinity
  Fields},'' \emph{IEEE Transactions on Pattern Analysis and Machine
  Intelligence}, 2019.

\bibitem{simon2017hand}
T.~Simon, H.~Joo, I.~Matthews, and Y.~Sheikh, ``{Hand Keypoint Detection in
  Single Images using Multiview Bootstrapping},'' in \emph{Conference on
  Computer Vision and Pattern Recognition}, 2017.

\bibitem{mmpose2020}
{MMPose Contributors}, ``{OpenMMLab Pose Estimation Toolbox and Benchmark},''
  \url{https://github.com/open-mmlab/mmpose}, 2020.

\bibitem{li2018crowdpose}
J.~Li, C.~Wang, H.~Zhu, Y.~Mao, H.-S. Fang, and C.~Lu, ``{Crowdpose: Efficient
  Crowded Scenes Pose Estimation and A New Benchmark},'' in \emph{Proceedings
  of the IEEE/CVF conference on computer vision and pattern recognition}, 2019,
  pp. 10\,863--10\,872.

\bibitem{xiu2018poseflow}
Y.~Xiu, J.~Li, H.~Wang, Y.~Fang, and C.~Lu, ``{Pose Flow}: {Efficient Online
  Pose Tracking},'' in \emph{British Machine Vision Conference}, 2018.

\bibitem{WangSCJDZLMTWLX19}
J.~Wang, K.~Sun, T.~Cheng, B.~Jiang, C.~Deng, Y.~Zhao, D.~Liu, Y.~Mu, M.~Tan,
  X.~Wang, W.~Liu, and B.~Xiao, ``Deep high-resolution representation learning
  for visual recognition,'' \emph{{IEEE Transactions on Pattern Analysis and
  Machine Intelligence}}, 2019.

\bibitem{wu2019detectron2}
Y.~Wu, A.~Kirillov, F.~Massa, W.-Y. Lo, and R.~Girshick, ``Detectron2,''
  \url{https://github.com/facebookresearch/detectron2}, 2019.

\bibitem{guler2018densepose}
R.~A. G{\"u}ler, N.~Neverova, and I.~Kokkinos, ``{Densepose: Dense Human Pose
  Estimation in the Wild},'' in \emph{Proceedings of the IEEE conference on
  computer vision and pattern recognition}, 2018, pp. 7297--7306.

\bibitem{mediapipe}
C.~Lugaresi, J.~Tang, H.~Nash, C.~McClanahan, E.~Uboweja, M.~Hays, F.~Zhang,
  C.-L. Chang, M.~Yong, J.~Lee, W.-T. Chang, W.~Hua, M.~Georg, and
  M.~Grundmann, ``{MediaPipe: A Framework for Perceiving and Processing
  Reality},'' in \emph{Third Workshop on Computer Vision for AR/VR at IEEE
  Computer Vision and Pattern Recognition}, 2019.

\bibitem{andriluka14cvpr}
M.~Andriluka, L.~Pishchulin, P.~Gehler, and B.~Schiele, ``{2D Human Pose
  Estimation: New Benchmark and State of the Art Analysis},'' in \emph{IEEE
  Conference on Computer Vision and Pattern Recognition (CVPR)}, June 2014.

\bibitem{mediapipe_hands}
\BIBentryALTinterwordspacing
F.~Zhang, V.~Bazarevsky, A.~Vakunov, A.~Tkachenka, G.~Sung, C.-L. Chang, and
  M.~Grundmann, ``{MediaPipe Hands: On-device Real-time Hand Tracking},'' 2020.
  [Online]. Available: \url{https://arxiv.org/abs/2006.10214}
\BIBentrySTDinterwordspacing

\bibitem{rim2020real}
B.~Rim, N.-J. Sung, J.~Ma, Y.-J. Choi, and M.~Hong, ``{Real-time Human Pose
  Estimation using RGB-D images and Deep Learning},'' \emph{Journal of Internet
  Computing and Services}, vol.~21, no.~3, pp. 113--121, 2020.

\bibitem{zimmermann20183d}
C.~Zimmermann, T.~Welschehold, C.~Dornhege, W.~Burgard, and T.~Brox, ``{3D
  Human Pose Estimation in RGBD Images for Robotic Task Learning},'' in
  \emph{2018 IEEE International Conference on Robotics and Automation
  (ICRA)}.\hskip 1em plus 0.5em minus 0.4em\relax IEEE, 2018, pp. 1986--1992.

\bibitem{pascual20223dhuman}
D.~Pascual-Hernández, N.~{Oyaga de Frutos}, I.~Mora-Jiménez, and J.~M.
  Cañas-Plaza, ``Efficient 3d human pose estimation from rgbd sensors,''
  \emph{Displays}, vol.~74, p. 102225, 2022.

\bibitem{fang2021_3dhuman}
Z.~Fang, A.~Wang, C.~Bu, and C.~Liu, ``3d human pose estimation using rgbd
  camera,'' in \emph{2021 IEEE International Conference on Computer Science,
  Electronic Information Engineering and Intelligent Control Technology (CEI)},
  2021, pp. 582--587.

\bibitem{nguyen2018}
D.~H.~P. Nguyen, M.~Hoffmann, A.~Roncone, U.~Pattacini, and G.~Metta, ``Compact
  real-time avoidance on a humanoid robot for human-robot interaction,'' in
  \emph{2018 13th ACM/IEEE International Conference on Human-Robot Interaction
  (HRI)}, 2018, pp. 416--424.

\bibitem{loper2015smpl}
M.~Loper, N.~Mahmood, J.~Romero, G.~Pons-Moll, and M.~J. Black, ``{SMPL: A
  Skinned Multi-Person Linear Model},'' \emph{ACM transactions on graphics
  (TOG)}, vol.~34, no.~6, pp. 1--16, 2015.

\bibitem{bogo2016keep}
F.~Bogo, A.~Kanazawa, C.~Lassner, P.~Gehler, J.~Romero, and M.~J. Black,
  ``{Keep it SMPL: Automatic Estimation of 3D Human Pose and Shape from a
  Single Image},'' in \emph{{European Conference on Computer Vision}}.\hskip
  1em plus 0.5em minus 0.4em\relax Springer, 2016, pp. 561--578.

\bibitem{kanazawa2018end}
A.~Kanazawa, M.~J. Black, D.~W. Jacobs, and J.~Malik, ``End-to-end recovery of
  human shape and pose,'' in \emph{Computer Vision and Pattern Recognition
  (CVPR)}, 2018.

\bibitem{pavlakos2019expressive}
G.~Pavlakos, V.~Choutas, N.~Ghorbani, T.~Bolkart, A.~A. Osman, D.~Tzionas, and
  M.~J. Black, ``{Expressive Body Capture: 3D Hands, Face, and Body from a
  Single Image},'' in \emph{Proceedings of the IEEE/CVF conference on computer
  vision and pattern recognition}, 2019, pp. 10\,975--10\,985.

\bibitem{omran2018neural}
M.~Omran, C.~Lassner, G.~Pons-Moll, P.~Gehler, and B.~Schiele, ``{Neural Body
  Fitting: Unifying Deep Learning and Model-Based Human Pose and Shape
  Estimation},'' in \emph{2018 international conference on 3D vision
  (3DV)}.\hskip 1em plus 0.5em minus 0.4em\relax IEEE, 2018, pp. 484--494.

\bibitem{pavlakos2018learning}
G.~Pavlakos, L.~Zhu, X.~Zhou, and K.~Daniilidis, ``{Learning to Estimate 3D
  Human Pose and Shape from a Single Color Image},'' in \emph{Proceedings of
  the IEEE conference on computer vision and pattern recognition}, 2018, pp.
  459--468.

\bibitem{docekal_thesis}
J.~Docekal, ``Close proximity human keypoint detection,'' Master's thesis, FEE
  CTU, 2022.

\bibitem{dataset}
\BIBentryALTinterwordspacing
J.~Docekal and M.~Hoffmann, 2022. [Online]. Available:
  \url{https://osf.io/qfkvt}
\BIBentrySTDinterwordspacing

\bibitem{icub}
G.~Metta, L.~Natale, F.~Nori, G.~Sandini, D.~Vernon, L.~Fadiga, C.~{von
  Hofsten}, K.~Rosander, M.~Lopes, J.~Santos-Victor, A.~Bernardino, and
  L.~Montesano, ``{The iCub humanoid robot: An open-systems platform for
  research in cognitive development},'' \emph{Neural Networks}, vol.~23, no.~8,
  pp. 1125--1134, 2010, social Cognition: From Babies to Robots.

\bibitem{oks}
M.~Ruggero~Ronchi and P.~Perona, ``{Benchmarking and Error Diagnosis in
  Multi-Instance Pose Estimation},'' in \emph{Proceedings of the IEEE
  international conference on computer vision}, 2017, pp. 369--378.

\bibitem{Roncone_IROS_2015}
A.~Roncone, M.~Hoffmann, U.~Pattacini, and G.~Metta, ``Learning peripersonal
  space representation through artificial skin for avoidance and reaching with
  whole body surface,'' in \emph{Intelligent Robots and Systems (IROS), 2015
  IEEE/RSJ International Conference on}, 2015, pp. 3366--3373.

\bibitem{Roncone2016}
A.~Roncone, M.~Hoffmann, U.~Pattacini, L.~Fadiga, and G.~Metta, ``Peripersonal
  space and margin of safety around the body: learning tactile-visual
  associations in a humanoid robot with artificial skin,'' \emph{{PLoS} {ONE}},
  vol.~11, no.~10, p. e0163713, 2016.

\bibitem{qualisys}
``{Qualisys Motion Capture System},''
  \url{https://www.qualisys.com/cameras/miqus/}, accessed: \today.

\bibitem{arun87}
K.~S. Arun, T.~S. Huang, and S.~D. Blostein, ``{Least-Squares Fitting of Two
  3-D Point Sets},'' \emph{IEEE Trans. on Pattern Analysis and Machine
  Intelligence}, no.~5, pp. 698--700, 1987.

\bibitem{lucci2020combining}
N.~Lucci, B.~Lacevic, A.~M. Zanchettin, and P.~Rocco, ``{Combining speed and
  separation monitoring with power and force limiting for safe collaborative
  robotics applications},'' \emph{{IEEE Robotics and Automation Letters}},
  vol.~{5}, no.~4, pp. {6121--6128}, 2020.

\end{thebibliography}

\end{document}